\documentclass[letterpaper, 10 pt, conference]{ieeeconf}  

\IEEEoverridecommandlockouts                              

\usepackage[T1]{fontenc}
\usepackage[utf8]{inputenc} 
\usepackage{todonotes}

\usepackage{graphics} 
\usepackage{amsmath} 
\DeclareMathOperator*{\argmax}{argmax}

\usepackage{multicol, blindtext}
\usepackage{xcolor}
\usepackage{booktabs}
\usepackage{caption}
\usepackage[normalem]{ulem} 

\title{\LARGE \bf
Dynamic Cloth Manipulation with Deep Reinforcement Learning 
}

\author{Rishabh Jangir, Guillem Alenyà, Carme Torras
\thanks{*This work has been supported by the ERC project Clothilde (ERC-2016-ADG-741930), the HuMoUR project TIN2017-90086-R (AEI/FEDER, UE) and by the AEI through the María de Maeztu Seal of Excellence to IRI (MDM-2016-0656). }
\thanks{The authors are  with Institut de Robòtica i Informàtica Industrial, CSIC-UPC,  
	Llorens i Artigas 4-6, 08028 Barcelona, Spain. {\tt  \{rjangir, galenya, torras\}@iri.upc.edu}}
}

\begin{document}

\maketitle
\thispagestyle{empty}
\pagestyle{empty}

\begin{abstract}
In this paper we present a Deep Reinforcement Learning approach to solve dynamic cloth manipulation tasks. Differing from the case of rigid objects, we stress that the followed trajectory (including speed and acceleration) has a decisive influence on the final state of cloth, which can greatly vary even if the positions reached by the grasped points are the same. We explore how goal positions for non-grasped points can be attained through learning adequate trajectories for the grasped points. Our approach uses few demonstrations to improve control policy learning, and a sparse reward approach to avoid engineering complex reward functions. Since perception of textiles is challenging, we also study different state representations to assess the minimum observation space required for learning to succeed. Finally, we compare different combinations of control policy encodings, demonstrations, and sparse reward learning techniques, and show that our proposed approach can learn dynamic cloth manipulation in an efficient way, i.e., using a reduced observation space, a few demonstrations, and a sparse reward.
\end{abstract}
\section{INTRODUCTION}
Day to day tasks of a household robotic assistant would involve manipulating deformable objects (cloth folding \cite{miller2012geometric}, bed making \cite{seita_bedmake_2019}, getting dressed \cite{gao2016iterative, Canal_icra18}). But the majority of state-of-the-art robotic manipulation work focuses on rigid objects.  Progress towards deformable object manipulation has been scarce, because manipulating deformable objects poses additional challenges over rigid objects. The shape of deformables varies largely along and between trajectories with the same end points and it is difficult to characterize due to their high-dimensional configuration spaces.\par

Deformable object manipulation can be achieved by static or dynamic manipulations. In dynamic manipulation, forces due to acceleration play a relevant role along with kinematics, static and quasi-static forces. Dynamic movement permits controlling non-grasped points as well, increasing in a way the dexterity of manipulation process at the expense of the increased complexity of underactuation. Recent research in deformable object manipulation (\cite{li2015folding, cusumano2011bringing, jimenez2012survey}) mostly considers only static manipulation tasks. The majority focus on explicit physics-based modeling of deformable behavior in simulation and they attempt to find an optimal trajectory to guarantee the desired outcome. Some studies have tried dynamic manipulations of flexible objects  (\cite{colome2018dimensionality, colome2015friction, ruggiero2018nonprehensile}) such as (dynamic) cloth folding, rope knotting, etc. with physics-based modeling (\cite{yamakawa2011}). A downside of these approaches and some other approaches that rely on visuomotor servoing (\cite{maitin2010cloth, osawa2007unfolding, bersch2011bimanual}) is the requirement of significant engineering specific to the manipulation task. Additionally, even while only considering static manipulation these models tend to be very sensitive to the deformation parameters of the object. It is then safe to say that these approaches would fail when considering dynamic manipulation tasks where the task performance heavily relies on the deformation parameters. \par

\begin{figure}[tb]
	\centerline{\includegraphics[totalheight=6.3cm]{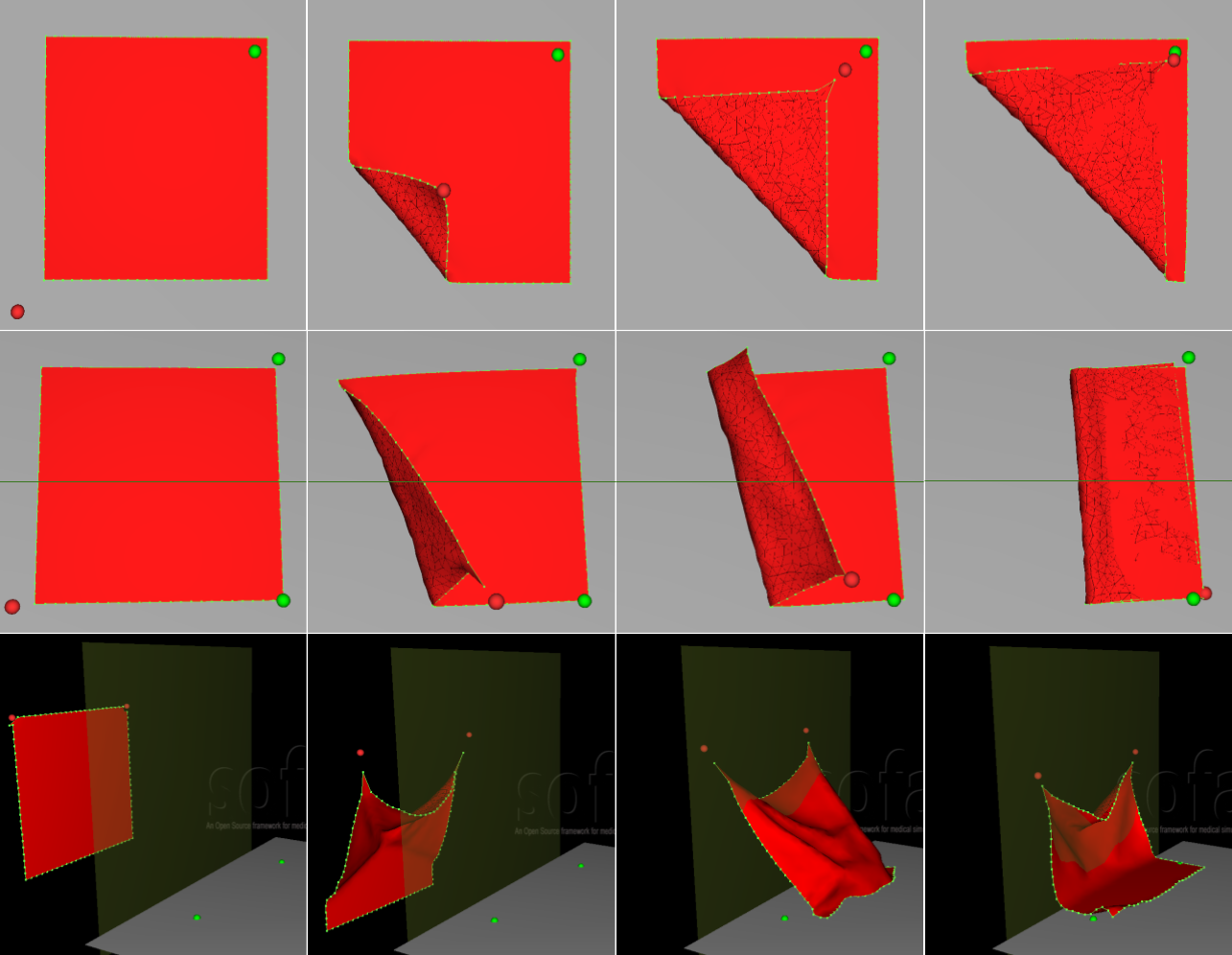}}
	\caption{Frames from rollouts of the learned policy are shown above for three tasks. The agent (red sphere) interacts with the cloth to move selected  vertice/s to their target positions. These are complex, under-actuated dynamic tasks where, to succeed, the agent needs to learn to successfully manipulate parts of cloth which are not in its direct control. }
	\label{fig:tasks_sequence}
	\vspace*{-3mm}
\end{figure}

Alternative approaches that employ trial-and-error-based machine learning methods (\cite{Corona_PR18,lee2015learning, balaguer2011combining, matas2018sim, hu2018three, ramisa2013finddd}) to map observations directly to actions have shown interesting results on robotic manipulation of deformable objects, but they restrict themselves to static manipulation as well. Regarding dynamic manipulation, \cite{kawaharazuka2019dynamic} developed a control method to realize a target state by calculating an optimized time-series joint torque command, but they mainly consider only a 2D actuation that allows only simple manipulation tasks and ~\cite{yang2016repeatable} considers only quasi-static movements.\par

Currently, training these model-free learning-based approaches is highly sample inefficient. Fortunately, simulations provide an efficient way to train control policies. However, accurate simulation of deformable objects is challenging. Previous approaches have investigated the usage of several simulators for deformable objects (\cite{coumans2016pybullet, lee2018dart}) but none of them have proved to be a benchmark for textile object simulation (like Mujoco \cite{todorov2012mujoco}  for rigid objects). 

Large configuration spaces can possibly be tackled by using RGB observation data. However, this would increase the computing cost by a large amount and, in addition, it is challenging to learn directly in the high-dimensional input state-space, which can also be redundant given the aim of the task. Moreover, current learning algorithms that use RGB information to solve cloth manipulation tasks \cite{matas2018sim} note the need of auxiliary inputs for the learning to succeed anyway. Nevertheless, they restrict themselves to solving static manipulations tasks only.

In order to address these problems, in this work we employ a model-free deep reinforcement learning (RL) method to learn dynamic manipulation of textile objects. Our method involves minimal task engineering as we take advantage of very few demonstrations and sparse rewards. This has been extensively studied in the context of rigid object manipulation \cite{quillen2018deep, nair2018overcoming, gu2017deep, peters2008reinforcement}, and a few studies on static deformable object manipulation \cite{matas2018sim, lu2015learning}, but to the best of our knowledge no study has previously investigated the applicability of deep RL methods to \textit{dynamic} deformable object manipulation tasks.  We use the SOFA simulator \cite{faure:hal-00681539} to define 3 textile manipulation tasks (Fig. \ref{fig:tasks_sequence}): one static manipulation task of folding a napkin diagonally, and two dynamic manipulation tasks of folding a napkin sideways and placing a napkin on a table. We provide a single sparse reward on task completion in all the tasks. Due to the challenging nature of the dynamic tasks being considered, we omit RGB information and directly work on the simplified clothing state consisting of a few distinguished points.

Through our experiments, we (a) demonstrate the basic difference between static and dynamic manipulations and show that the simulator is capable of capturing these differences. (b) Investigate the trade-off between complexity and effectiveness to find an accurate textile state representation for our tasks. (c) Show that our method can learn valid dynamic manipulation behaviors using a low-fidelity simulation platform and test different combinations of the components of the algorithm to show their individual effect.


\section{BACKGROUND} \label{background}
We consider a standard RL problem where an agent interacts with an environment according to a policy in order to maximize rewards over discrete timesteps. The framework considers partially observable environments that are modeled as continuous state partially observable Markov decision processes (POMDPs) defined as a tuple (\(S\),\(A\),\(P\),\(r\), \(O\), \(\rho_{0}\),\(\gamma\)), where \(S\) is a set of full states of the environment, \(A\) is a set of continuous actions, \(P: S\times A\times S\to R\)  is the transition probability distribution, \(r:S\times A \to R\) is the reward function, \(\rho_{0}\) is the initial state distribution, and \(\gamma \in (0,1]\)  is the discount factor. The decision process is partially observable and the agent receives observations \(o\) from the set of observations \(O\).

The goal of the agent is to maximize the multi-step return \(R_t=\sum_{t'=t}^T\gamma^{t'-t} r_{t'}\), where \(T\) is the fixed horizon of each episode. The objective during learning is to find an optimal policy \(\pi^*:O \to A\) that maximizes the expected return of the agent
 \(J(\pi)\)
\begin{equation}
\pi^*= \argmax_{\pi} J(\pi).
\end{equation}
lternatively, the expected return upon taking an action \(a_t\) in state \(s_t\) can be measured by a \(Q\) function \(Q(s_t,a_t)=E[R_t|s_t,a_t]\). In terms of \(Q\) function the objective can be written as, 
\begin{equation}
\pi^*= \argmax_{\pi} Q(s_t,\pi(o_t)).
\end{equation}
The transition probability distribution \(P\) determines the consequences of the agent's actions and is dependent on the dynamics of the environment. The dynamics is therefore of crucial importance, as it determines the behaviors that can be realized. Deformable objects introduce an additional component in the environment dynamics which is lacking for rigid objects, making it more difficult to learn valid behaviors without direct access to the environment dynamics.

\subsection{Deep Deterministic Policy Gradients}
DDPG \cite{lillicrap2015continuous} is a policy gradient algorithm for learning control policies in continuous action domain. It uses off-policy data and the Bellman equation to concurrently learn the Q-function and a policy. It uses an actor neural network (policy network), parameterized by a set of parameters \(\theta^{\pi}\), that maps observations to actions \(\pi(\theta):O \to A\) and tries to maximize \(Q(s_t,\pi_{\theta}(o_t))\) at each time-step \(t\). However, the \(Q\) function is not known and DDPG employs a critic neural network, parameterized by parameters \(\theta^Q\), to estimate \(Q\) values of actions at each time-step \(t\).

During training, the agent acts in the environment according to noisy policy \(a_t=\pi(o_t)+N(0,\sigma)\). The Gaussian noise facilitates exploration. Each transition the agent generates is stored in a replay buffer \(R\) from where it is sampled in batches of \(N\) tuples to train the networks. Sampling from a replay buffer stabilizes training by removing temporal correlations and therefore reduces the changes in the distributions the networks are trying to learn. DDPG also employs target networks \(Q^*\) and \(\pi^*\) to reduce the risk of \(Q\) value estimates oscillating or diverging due to the recursive \(Q\) value definition in the Bellman equation.
\subsection{Universal Policy}
A universal policy \cite{schaul2015universal} is a simple extension where the goal \(g\in G\) is provided as an additional input to the policy \(\pi(a|o,g)\). The reward is then also dispensed according to the goal \(r(s_t,a_t,g)\). In our framework, a different goal will be randomly sampled at the start of each episode, and held fixed over the course of the episode. For the cloth manipulation tasks, the goal specifies target location for the selected vertices as discussed later.
\subsection{Hindsight Experience Replay}
Learning from a sparse binary reward is known to be challenging for most modern RL algorithms. We will therefore leverage a recent innovation, Hindsight Experience Replay (HER) \cite{andrychowicz2017hindsight}, to train policies using sparse rewards. In this work, we consider sparse binary rewards of the form, \(r(s,g) = 0\) if \(g\) is satisfied in \(s\), and \(r(s,g) = -1\) otherwise.
Consider an episode with trajectory \(\tau \in (s_0,a_0,...,a_{T-1},s_{T})\), where the goal \(g\) was not satisfied over the course of the trajectory. Since the goal was not satisfied, the reward will be -1 at every timestep, providing little information to the agent. 

But suppose that we are provided with a mapping \(m:S\to G\), that maps a state to the corresponding goal satisfied in the given state. For example, \(m(s_{T})=g'\) represents the goal that is satisfied in the final state of the trajectory. Rewards are then recomputed under the new goal \(g'\). The trajectory will be successful under the new goal as \(g'\) is satisfied in \(s\). By replaying past experiences with HER, the agent can be trained with more successful examples than those available in the original recorded trajectories. We have considered replaying 4 goals from the future states of a trajectory, but HER is also amenable to other replay strategies.
\subsection{Learning from demonstrations}
Exploration in continuous state-space environments is another existing challenge for deep RL algorithms. Moreover, the exploration problem is magnified when we consider deformable object environments due to their large configuration spaces. We combine DDPG algorithm with several practical extensions as introduced in (\cite{nair2018overcoming, vevcerik2017leveraging}), namely an additional buffer to store demonstration data, a modified loss that uses behavior cloning loss as an auxiliary loss, and a filter on the \(Q\) values which accounts for sub-optimal demonstrations.

\begin{figure*}[h]
	\begin{multicols}{3}
		\captionsetup{labelformat=empty}
		\includegraphics[width=1\linewidth]{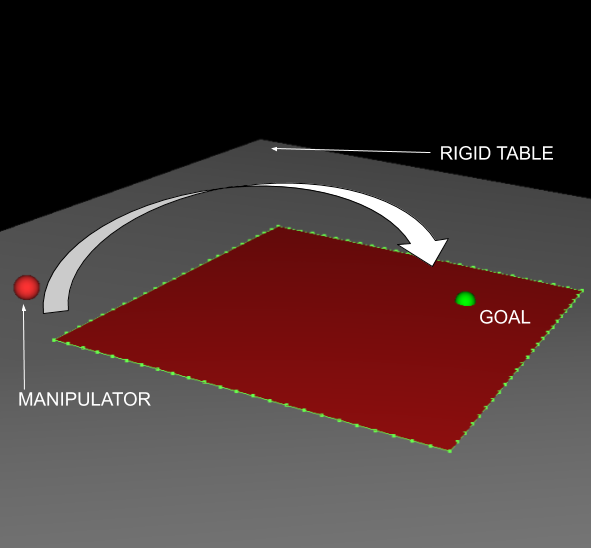}
		\caption{Diagonal Folding (\(T\)=200 steps)}\addtocounter{figure}{-1}\par
		\includegraphics[width=1\linewidth]{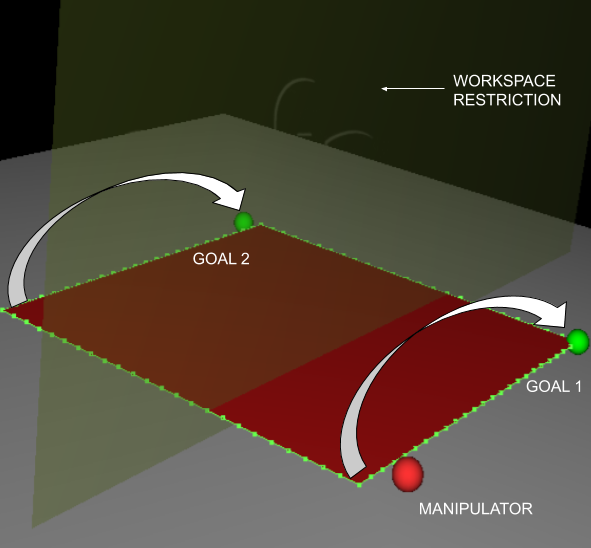}
		\caption{Sideways Folding (\(T\)=300 steps)}\addtocounter{figure}{-1}\par
		\includegraphics[width=1\linewidth]{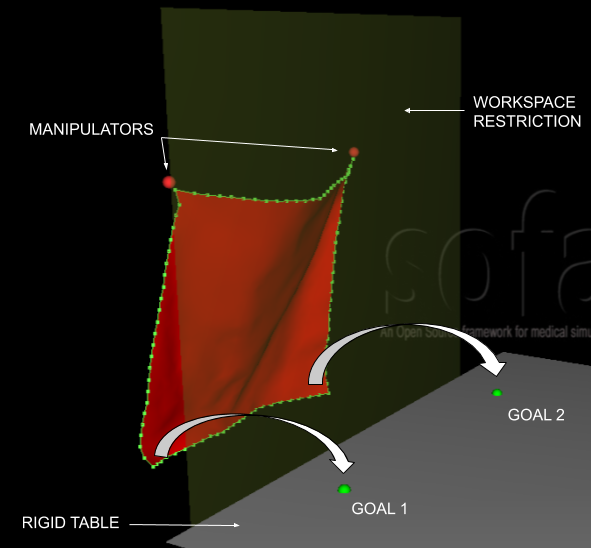}
		\caption{Placing on a table (\(T\)=500 steps)}\addtocounter{figure}{-1}
	\end{multicols}
	\vspace*{-3mm}
	\caption{Task definition diagrams are shown for each task. The goal state is reached when selected vertice/s are within a certain threshold of their target positions. The target spheres (green) do not interact with the environment. In the last two tasks the agent's workspace is restricted as shown by a green plane and thus dynamic manipulations are required to successfully reach the goal state in these tasks.}
	\label{fig:tasks_define}
	\vspace*{-3mm}
\end{figure*} 

\section{METHOD} \label{method}

\subsection{Simulation}
Development towards finding effective deformable object simulators has been scarce. This is because of the difficulty of efficiently simulating deformations and the large diversity observed in case of deformable objects. Although there have been efforts towards using existing simulators that support deformable objects for deep RL research (\cite{coumans2016pybullet,lee2018dart}) there have been no benchmark simulations or tasks. An important case of deformable objects is textile/clothing. Matas et al.\cite{matas2018sim} used PyBullet \cite{coumans2016pybullet} for simulating towel manipulation tasks, but they reported issues with the simulator codebase such as instability as it only implements some rudimentary and experimental functionality. Instead, for this work we use the SOFA simulation framework~\cite{faure:hal-00681539}. SOFA provides us with more realistic and intricate deformations of the clothing. In comparison with PyBullet we found that SOFA  was much more stable and flexible. Although the simulator still does not support self-collisions in the cloth, it provides functionalities to define the physical properties associated with the kind of deformation required.

\subsection{Cloth manipulation environments}
We designed and implemented 3 environments using OpenAI gym ~\cite{brockman2016openai} API for solving manipulation tasks involving a square cloth of fixed shape and size placed on or around a rigid table. The cloth is modeled as a mesh of triangles joined together by their vertices. We call these vertices nodes in the cloth mesh, mass of the cloth is uniformly spread among all nodes. A rigid ball is used as a manipulator which can attach itself to a node on the cloth. The central aim of this work is to study dynamic manipulation of textiles, thus we bypass the need to model a physical grasp in the simulation. Instead, we use a fake grasp implemented as a binary point grasp to manipulate the textile. Creation of the grasp constraint is subject to the manipulator being in close proximity to a cloth node. Other forces acting on the clothing apart from the manipulator are gravity and interaction forces with the table such as friction.

\subsection{Tasks}

Fig. \ref{fig:tasks_define} shows the textile manipulation tasks we consider in the environments defined above. Each task runs for a fixed number of simulation steps \(T\) (different for different tasks) before it automatically gets reset to the initial state. At the start of each episode, the manipulator is initialized to a default pose near the vertex to be manipulated, and the cloth is placed randomly within a fixed bound on the table or around it depending on the task. There are no deformations/wrinkles in the cloth at the start of an episode. Goal state \(g\) varies for each task depending on the aim and definition of tasks which we discuss later. For dynamic manipulation tasks the motion of manipulator is constrained within a predefined workspace in order to demonstrate the emergence of dynamic behaviors. The 3 environments are,\par
\textbf{Diagonal (Static) Folding:} This task involves diagonally folding a piece of cloth lying on a table. Goal state is achieved if the vertex being manipulated is within a threshold distance (\(\delta\)=10 units) of the goal location. The goal location is randomly sampled along the diagonal in close proximity to the opposite vertex. No restrictions on the workspace of the manipulator in this task.\par
\textbf{Sideways (Dynamic) Folding:} This task involves sideways folding a piece of cloth lying on a table. Goal state is achieved if both pairs of adjacent vertices are within a threshold distance (\(\delta\)=10 units) of goal location. The goal locations are randomly sampled close to the destination vertices. The manipulator workspace is constrained in such a way that it cannot reach the other half of the cloth. The only way to manipulate the vertex in the other half of the cloth then is to rely on the fabric connection between the vertices and swing the other vertex to its goal location.\par
\textbf{(Dynamic) Placing on a table}: This task involves partially placing a piece of cloth on a table. The cloth hangs parallel to an edge of the table outside the table space. Two vertices of the cloth are grasped by two rigid ball manipulators at all times and the other two vertices are freely hanging. Goal state is achieved when both hanging vertices are within a threshold distance (\(\delta\)=20 units) of goal locations. Two goals are randomly sampled such that the line connecting them is parallel to the edge of the table. Similar actions are applied to both the manipulator and action space for this task is 3D. The manipulator workspace is constrained and thus it  cannot go beyond the edge of the table. The only way to reach the goal state is to swing the cloth such that the hanging vertices land on the goal locations.\par

\subsection{State and Action}
The state is represented using positions and velocities of selected nodes on the cloth, position and velocity of the manipulator, position of the goal/s, as well as grasp state of the manipulator. The combined features result in a state space ranging from 34D to 85D depending on the number of nodes selected on the cloth and the task. Actions from the policy specify target velocities for the manipulator in x, y and z directions as well as a  boolean gripping action. Manipulator rotation is not necessary for the tasks and is therefore kept fixed. This yields a 4D action space for the first two tasks and 3D action space for the last task where gripping action is not used.

\subsection{Learning behaviors from demonstrations algorithm}
During initial exploration we found that DDPG alone was not successful in solving any of the proposed tasks. We suspected the failure was due to the additional complexity that deformable objects bring to the environment with their infinite configuration spaces. So we investigated possible improvements (\cite{nair2018overcoming,vevcerik2017leveraging}) alongside vanilla DDPG. This led us towards employing demonstrations to aid the agent in faster exploration. Firstly, an additional buffer  \(R_D\) was initialized to contain demonstration data, at training time we draw an extra \(N_D\) samples from this demonstration buffer along with environment interaction data from main buffer in batches. Secondly, behavior cloning loss \(L_{BC}\) was used as an auxiliary loss to train the actor network only on the demonstration samples in the data defined as,
\begin{equation}
L_{BC}=  \sum_{i=1}^{N_D}(\pi(o_i,Q_{\pi})-a_i)^2 \mathbf{1}_{Q(s_i,a_i)>Q(s_i,\pi(o_i))}.
\label{eqn:lbc}
\end{equation}
In DDPG, a critic network is trained to predict \(Q\) values of the actions taken by the actor. The final part in Equation \ref{eqn:lbc} corresponds to a filter on the \(Q\) value update, that allows update only if the demonstrated action had a better \(Q\) value than the action output by the policy network. This results in the following loss functions, \(L_{\pi}\) for the actor 
\begin{equation}
L_{\pi}=\lambda_1 \nabla_{\theta_{\pi}}J-\lambda_2 \nabla_{\theta_{\pi}} L_{BC}
\end{equation}
where \(\lambda_1\) and \(\lambda_2\) are hyper-parameters corresponding to the weight shared between the two loss terms in \(L_{\pi}\), and \(L_Q\) for critic  network,
\begin{equation}
L_{Q}=\frac{1}{N} \sum_{i}\left((r_i+\gamma Q(s_{i+1},\pi(o_{i+1}))-Q(s_i,a_i|\theta_Q)\right)^2.
\end{equation}
Training the agent with demonstrations aided in exploration and it was able to solve the Diagonally folding task, still it failed in the two tasks involving dynamic manipulation. This can be attributed to the complex nature of the dynamic tasks and only sparse rewards being provided upon success. We further employed HER to handle the sparse nature of our data. HER augments the original training data with more successful examples by replaying trajectories with modified goals. HER only works with stationary goals and thus we designed tasks in such a way that the goal locations are fixed throughout the episode. We study the performance of individual components of the algorithm in Section \ref{experiments}.

\subsection{Capturing dynamics in a simulation}
The basic difference in dynamic manipulation as compared to static one is the manipulator being able to control non-grasped vertices. For example, in the Sideways Folding task the manipulator relies on the connection between grasped vertex (no.3) and the vertex to be manipulated (no.2) to successfully achieve the goal state. Effectively, the manipulator is being able to control parts of the object outside its workspace. The manipulating agent's ability should thus encompass learning deformation behavior as well as the effects of  manipulator's velocity and path on the outcome of the task. Consequently, the simulator must be able to capture these differences in our chosen tasks. In Section \ref{experiments} we present experimental results to show the relevance of manipulator trajectory, velocity and acceleration in the case of dynamic manipulation tasks and the simulator's ability to differentiate between task-relevant motion features.

\subsection{Generating demonstrations}
The demonstrations for all the tasks are generated by a hard-coded python script. The actuator follows points in the cloth reference frame that are pre-selected by a human in just one instance of the tasks. A uniform Gaussian noise of 10\% is added to introduce randomness in the demonstration data. The generated demonstrations are imperfect and we use 20 episodes of demonstration data to train our agent.

\section{EXPERIMENTS} \label{experiments}
We have performed three sets of experiments to study the effectiveness of reinforcement learning in solving dynamic deformable object manipulation tasks. The first set aims to unravel the basic difference between dynamic and static manipulation. The second is intended to assess the influence of state representation on performance. And the third tries to solve a difficult dynamic manipulation problem on deformable environments that had not yet been studied in the RL community, also we study the effect of individual components of the algorithm.

\subsection{Capturing dynamics in a simulation}
In order to assess the importance of task-relevant motion features, we introduce randomization in the speed and the trajectory of manipulator while keeping initial and final positions intact in our hard-coded demonstration generation script. In Table \ref{tab:capturing_dynamics} we report success rates under different randomization for all three tasks averaged over 3 epochs of 100 episodes each. \par
Success rates for dynamic tasks drop significantly under all types of randomization whereas negligible effect on the success rate of the static task is observed. This can be attributed to the fact that in the Diagonal Folding task the vertex in consideration is always grasped by the manipulator throughout the trajectory, so only final position of the manipulator matters for task success.  Thus we can conclude that, in dynamic tasks speed and trajectory play an important role in deciding task success. Additionally the results suggest that the simulator is capable of capturing the dynamic information in the proposed tasks.

\begin{table}[tb] 
	\centering
	\begin{tabular}{|c|c|c|c|}
		\hline
		& Diagonal & Sideways & Placing\\
		Randomization	& Folding & Folding & on table\\
		& (Static) & (Dynamic) & (Dynamic)\\
		\hline
		None	& $1.0\pm 0.00$ & $0.99\pm 0.01$ & $0.52\pm 0.04$ \\
		\hline
		Speed	& $0.99\pm 0.01$ & $0.41\pm 0.07$ & $0.41\pm 0.06$ \\
		\hline
		Trajectory	& $1.0\pm 0.00$ & $0.12\pm 0.03$ & $0.09\pm 0.05$\\
		\hline
		Speed+Trajectory & $1.0\pm 0.0$ & $0.08\pm 0.04$ & $0.14\pm 0.02$ \\
		\hline
	\end{tabular}
	\caption{Success rate comparisons for all three tasks under different randomization. Note that randomization has negligible effect on success rate for static tasks.}
	\label{tab:capturing_dynamics}
	\vspace*{-5mm}
\end{table}

\begin{figure}[tb]
	\centerline{\includegraphics[width=0.8\linewidth]{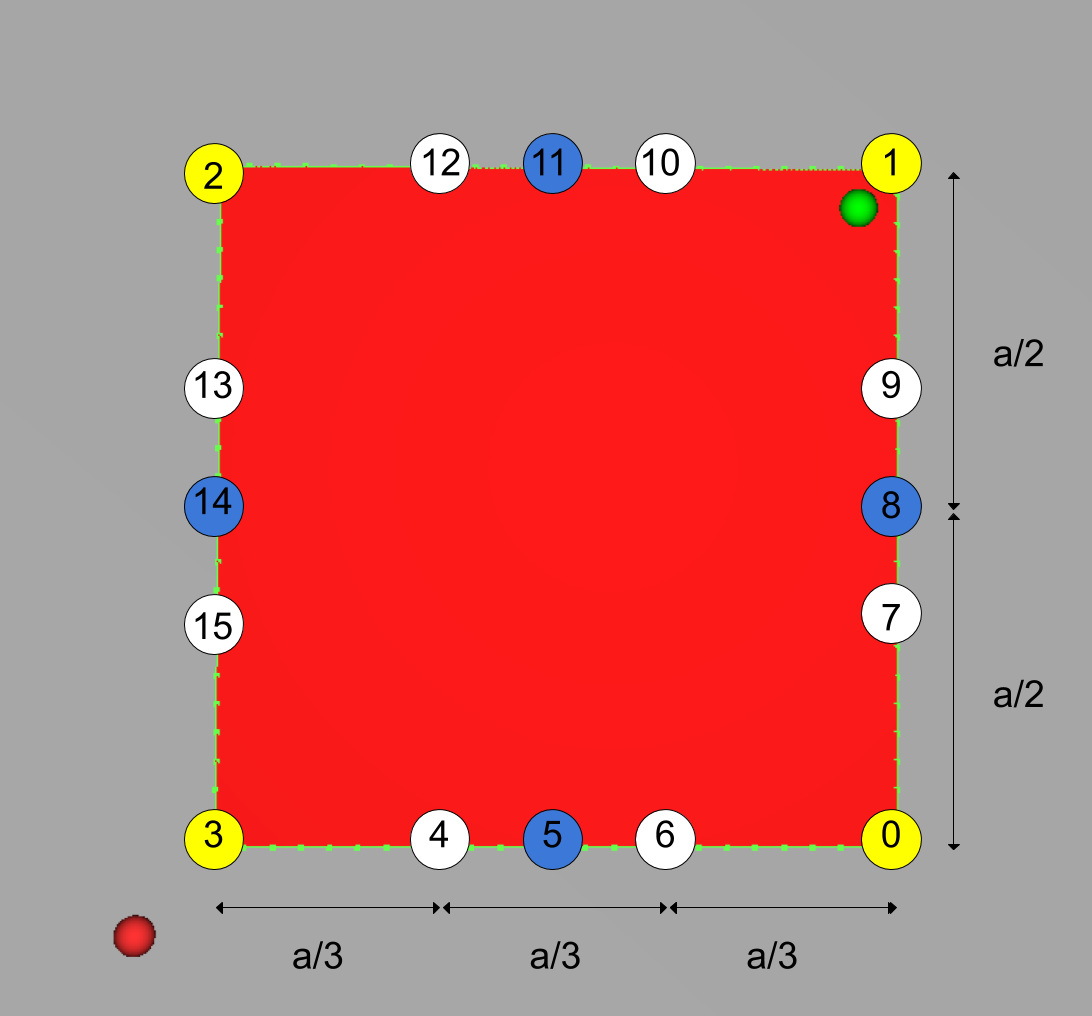}}
	\caption{Positions of selected points on the cloth. The 3 input observation
spaces considered are: (1) 4 yellow points, (2) 4 yellow and
4 blue points, (3) 4 yellow and 8 white points.}
	\label{fig:tasks_vertice_positions}
	\vspace*{-5mm}
\end{figure}

\begin{figure*}[h]
	\begin{multicols}{3}
		\captionsetup{labelformat=empty}
		\includegraphics[width=1.0\linewidth]{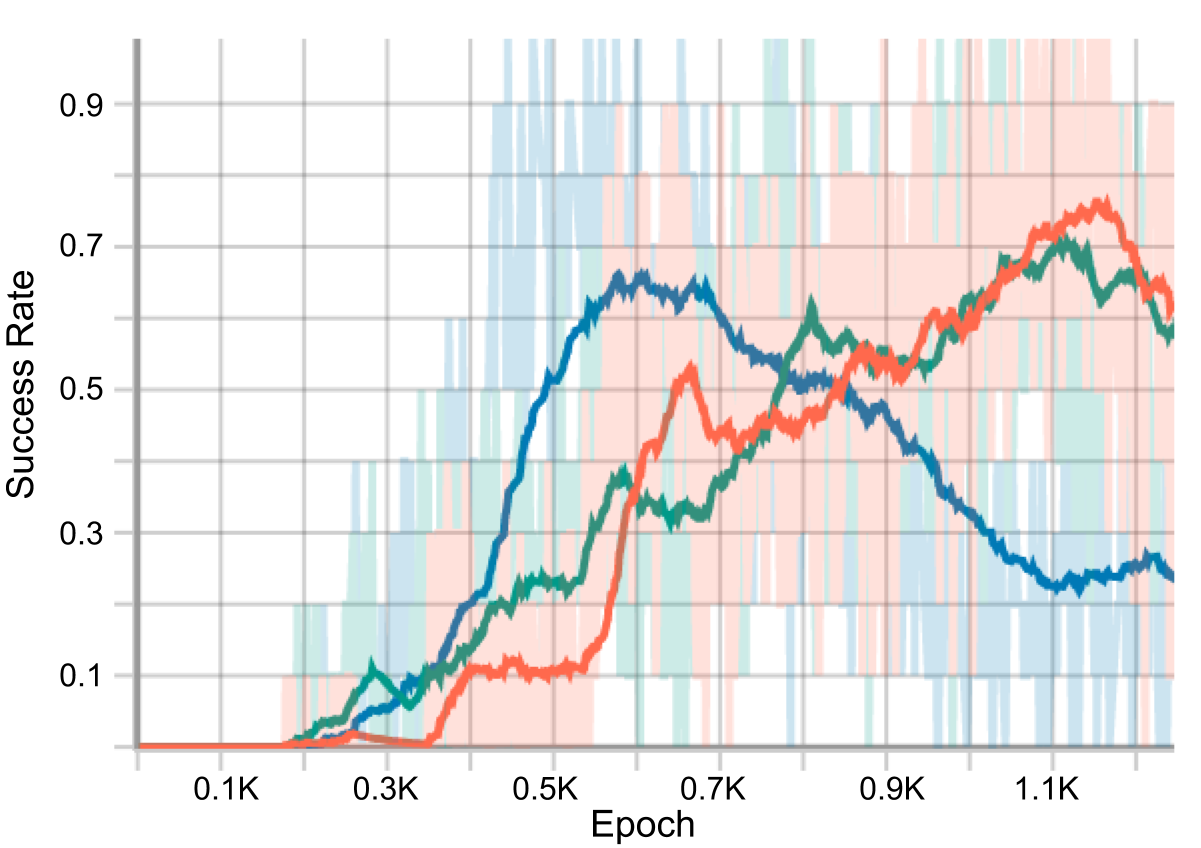}
		\caption{Diagonal Folding}\addtocounter{figure}{-1}\par
		\includegraphics[width=1.0\linewidth]{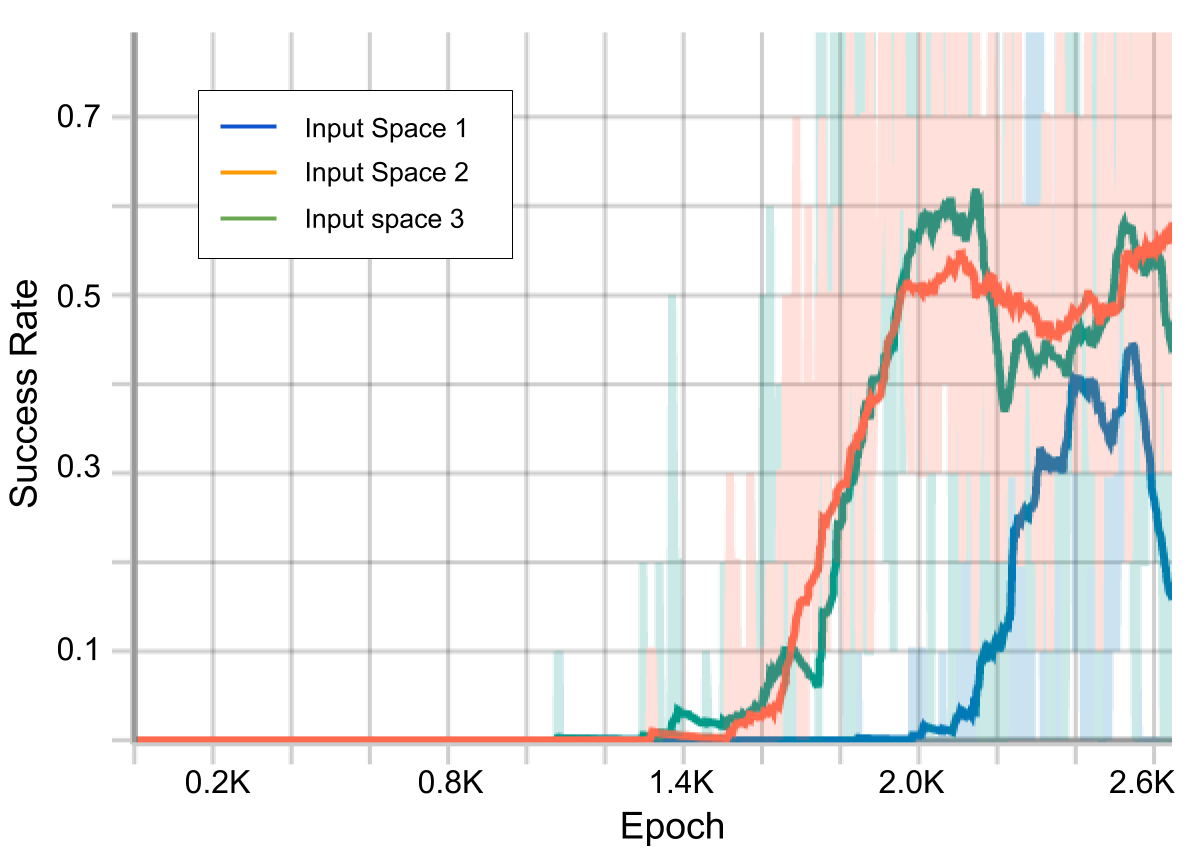}
		\caption{Sideways Folding}\addtocounter{figure}{-1}\par
		\includegraphics[width=1.0\linewidth]{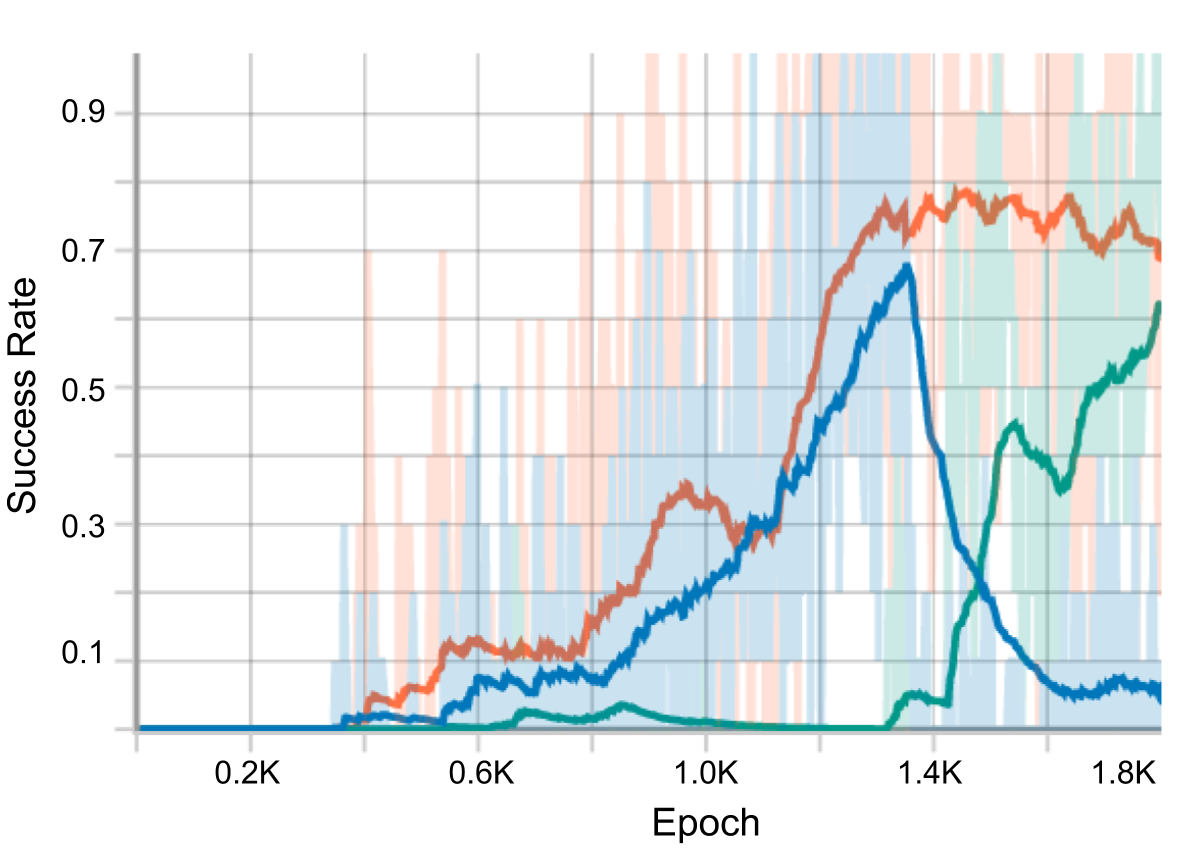}
		\caption{Placing on a table}\addtocounter{figure}{-1}
		
	\end{multicols}
	\vspace*{-3mm}
	\caption{Learning curves comparing performance of different observation space inputs. We use 3 random seeds per method and report average success rate. The median of the runs is shown in bold and each training run is plotted in a lighter color.}
	\label{fig:tasks_results_vertices}
	\vspace*{-2mm}
\end{figure*}

\begin{figure*}[h]
	\begin{multicols}{3}
		\captionsetup{labelformat=empty}
		\includegraphics[width=1.0\linewidth]{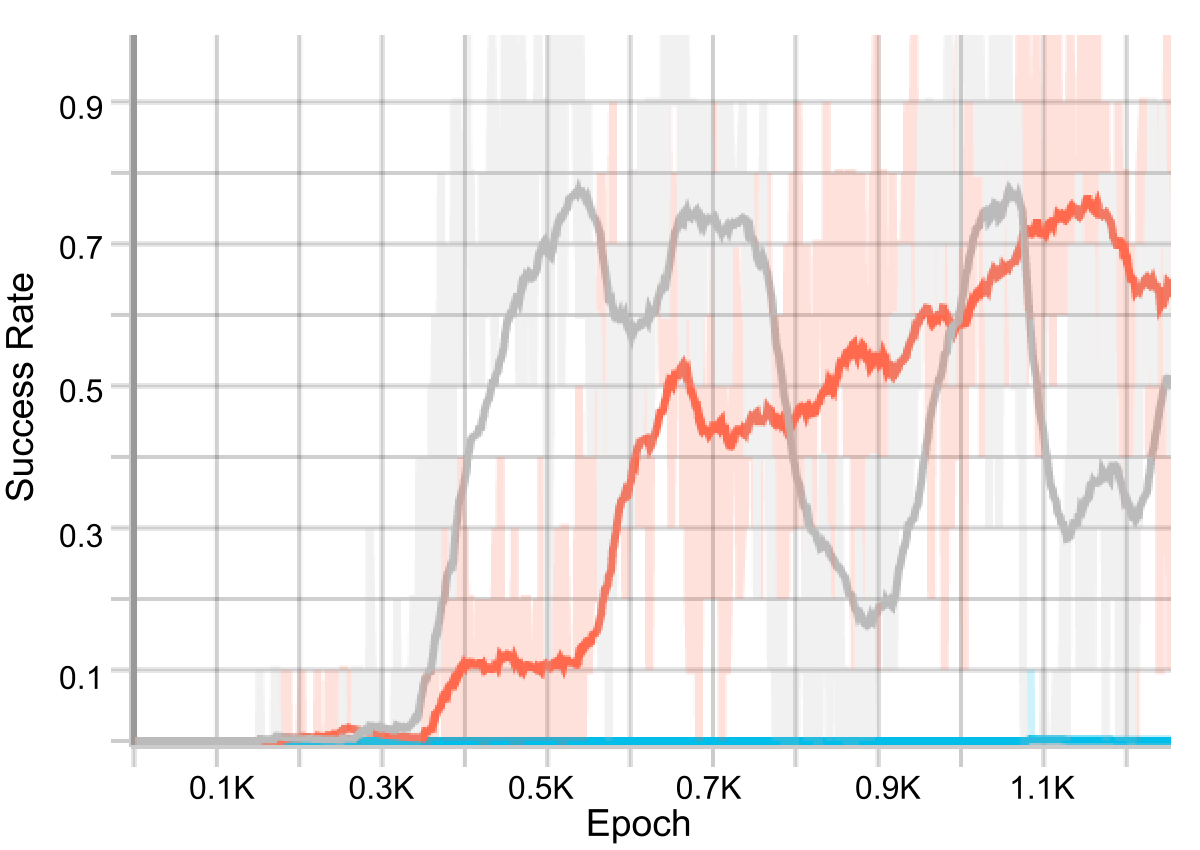}
		\caption{Diagonal Folding}\addtocounter{figure}{-1}\par
		\includegraphics[width=1.0\linewidth]{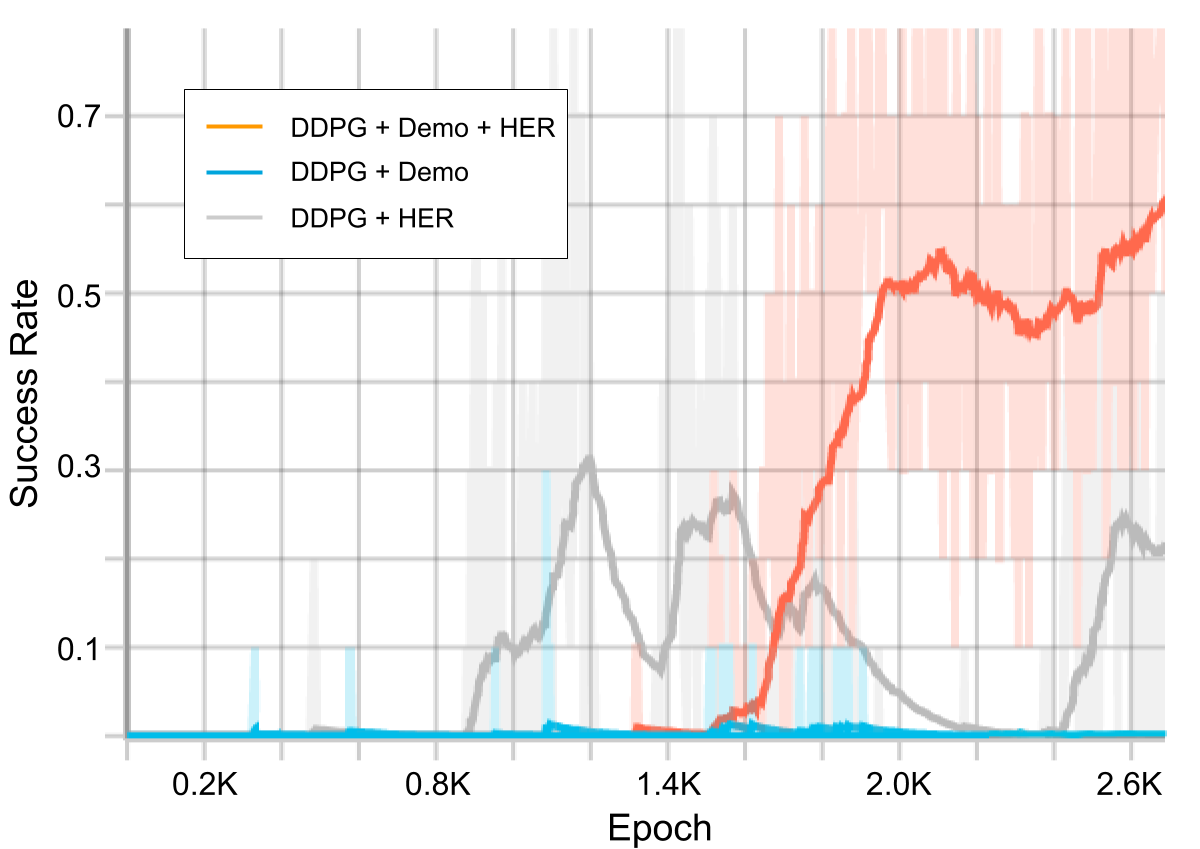}
		\caption{Sideways Folding}\addtocounter{figure}{-1}\par
		\includegraphics[width=1.0\linewidth]{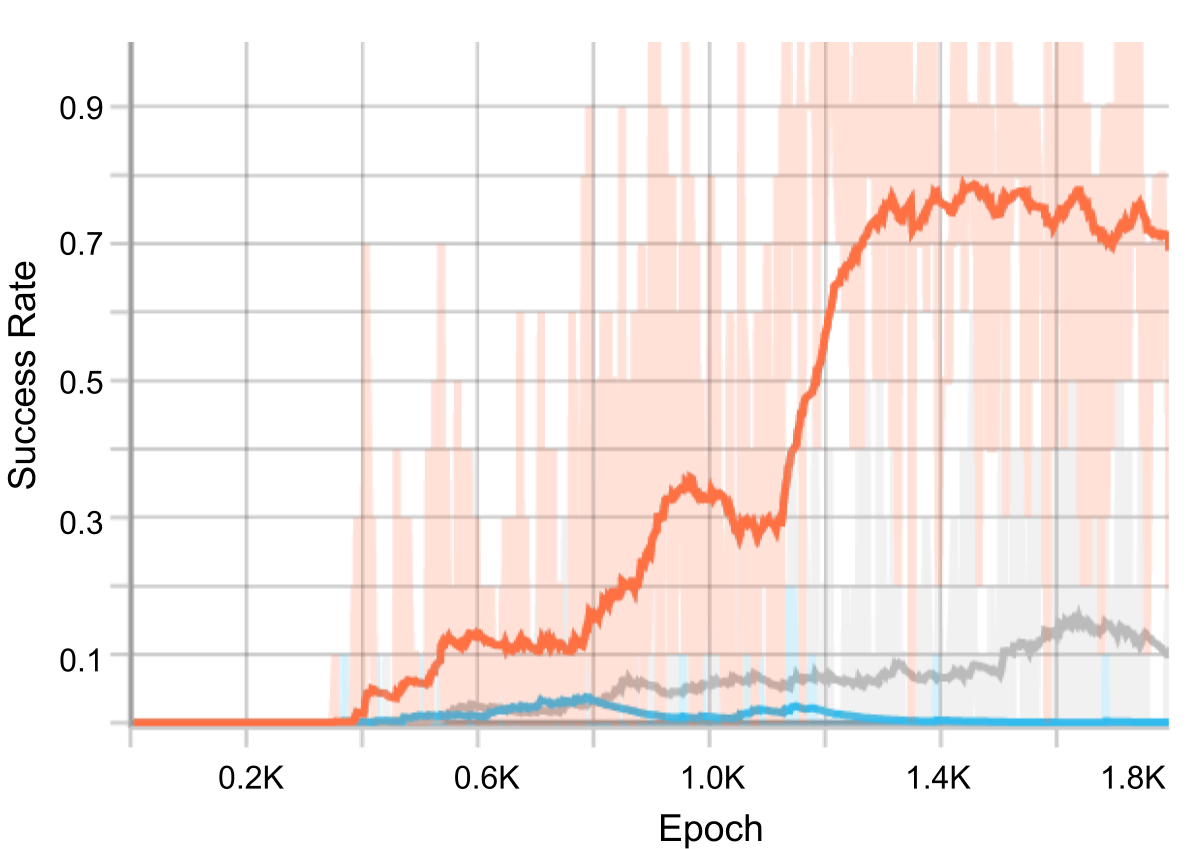}
		\caption{Placing on a table}\addtocounter{figure}{-1}
	\end{multicols}
	\vspace*{-1mm}
	\caption{Learning curves comparing training with and without demonstrations and HER along with DDPG. We use 3 random seeds per method and report average success rate. Note that the "DDPG+Demo" method without HER is always at around 0\% success rate. As the difficulty of the task increases demonstrations become essential for good performance.}
	\label{fig:tasks_results_her}
	\vspace*{-3mm}
\end{figure*}

\subsection{Textile state representation}
To investigate the effectiveness of different textile state representations, we compare agents trained with 3 different configurations of observation spaces as defined in Fig. \ref{fig:tasks_vertice_positions}. The main difference between the three alternatives is the size of observation space which increases from 4 to 12. We train our best performing "DDPG+Demo+HER" agent  (as defined in Sec.~\ref{sec:dynamicManipulation}) with 3 random seeds for each configuration. Learning curves comparing the performance of each input space can be seen in Fig. \ref{fig:tasks_results_vertices}.

We observe that 4 points is an insufficient state representation for solving the proposed tasks. A sudden drop in performance can be observed for this case which is possibly due to the algorithm learning a sub-optimal policy given how little deformation information is captured by only 4 vertex points. All the tasks show an improvement in performance (higher success rate, stable learning curves) as we increase the observation space to 8 points.  Further increment in the observation space size does not affect the performance in Diagonal Folding and Sideways Folding tasks as can be seen by the overlapping orange and green curves. Whereas for the same change, decrement in performance for Placing on table task is observed. Note that, increasing the observation space size increases training time of the algorithm. Therefore we can conclude 8 points are the best choice of observation space for the tasks considered in this work.

\subsection{Dynamic manipulation of deformable objects} \label{sec:dynamicManipulation}
We train 3 agents on our proposed tasks in simulation and do performance comparisons to study the importance of individual components. All 3 agents use DDPG as their base algorithm with 4 layered actor and critic networks. Also, they share the same configuration of all hyperparameters except the replay buffer data which also includes demonstration data (\(N_D\)=1/8 \(\times\) Total sampled data) and modified actor loss \(L_{\pi}\) to \(\lambda_1=0.001, \lambda_2=0.0078\) from \(\lambda_1=1.0, \lambda_2=0.0\) in the case of learning from demonstrations. We consider HER and training with demonstrations (Demo) as improvements and thus the 3 agent combinations are "DDPG+Demo+HER", "DDPG+HER" and "DDPG+Demo". We use 8 points for input space as determined in the previous experiment. At training time, each epoch consists of 20 episodes of training followed by 10 episodes of policy testing. The success rate at each epoch is measured as the average over 10 test episodes. We train each agent with 3 random seeds. Qualitative performance of the learned policies for the three tasks can be seen in the attached video where different executions of the learned policy are demonstrated and in Fig.~\ref{fig:tasks_sequence}, whereas learning curves comparing the performance of each agent can be seen in Fig.~\ref{fig:tasks_results_her}.

The "DDPG+Demo+HER" agent is able to solve all of the proposed tasks and successfully learn valid dynamic manipulation behaviors for the Sideways Folding and Placing on a table tasks. Looking at the learning curves, we observe that training without HER is unable to solve any of the tasks. Moreover, training without demonstrations is able to achieve comparable success only for the static task but the learning is highly unstable. This can be attributed to insufficient exploration and reaching sub-optimal minima without the presence of demonstrations to guide the agent towards good behaviors. An example of a sub-optimal behavior we observed while training for the diagonal folding task is the agent learning to grasp the cloth around midpoint of the line joining goal position and vertex and trying to swing the vertex to goal. The importance of demonstrations is magnified in the case of dynamic tasks where performance of "DDPG+HER" agent is significantly lower. But demonstrations are of little use for the above tasks when training without HER as observed with the "DDPG+Demo" agent. 

\section{CONCLUSIONS}

Building upon recent work in end-to-end learning for rigid object manipulation, we extended the ideas to dynamic cloth manipulation. We demonstrated the importance of speed and trajectory in the case of dynamic manipulations and investigated the effectiveness of different textile state representations. By restricting the manipulator workspace we showed the emergence of dynamic manipulation behaviors and successfully solved 3 long horizon tasks: Diagonal Folding, Sideways Folding and Placing on a table, using a deep RL method which bypasses the need to explicitly model cloth behavior and does not require reward shaping to converge. Also, we studied the importance of individual components of our algorithm which highlighted the importance of HER and demonstrations. We think these preliminary results hold promise that \textit{versatile} cloth manipulation by robots is within reach of current machine learning techniques. In the future, we plan to extend our work to real robots by training image to action policies in simulation and transferring the learned policy with domain randomization.

\addtolength{\textheight}{-2.0cm}   





\bibliographystyle{IEEEtran}
\bibliography{biblio}




\end{document}